\documentclass{article}

\usepackage{arxiv}

\usepackage[utf8]{inputenc} 
\usepackage[T1]{fontenc}    
\usepackage{hyperref}       
\usepackage{url}            
\usepackage{booktabs}       
\usepackage{amsfonts}       
\usepackage{nicefrac}       
\usepackage{microtype}      
\usepackage{lipsum}
\usepackage{float} 
\usepackage{graphicx}
\usepackage{natbib}
\usepackage{amsmath}
\usepackage{authblk}
\graphicspath{ {./images/} }
\usepackage{color,soul}

\newcommand{\R}{\mathbb{R}}

\title{Learning When to Stop: Adaptive Latent Reasoning via Reinforcement Learning}

\author[1,2]{
 Alex Ning\thanks{Corresponding author: rnx2bc@virginia.edu}\,
}
\author[1]{
 Yen-Ling Kuo
}
\author[2]{
 Gabe Gomes
}

\affil[1]{University of Virginia\\}
\affil[2]{Carnegie Mellon University\\}

\begin{document}
\maketitle
\begin{abstract}

Latent reasoning represents a new development in Transformer language models that has shown potential in compressing reasoning lengths compared to chain-of-thought reasoning. By directly passing the information-rich previous final latent state into the next sequence, latent reasoning removes the restriction to human language tokens as the medium for reasoning. We develop \textit{adaptive}-length latent reasoning models and introduce a post-SFT reinforcement-learning methodology to optimize latent reasoning length by minimizing reasoning length while maintaining accuracy. This, in turn, further reduces compute usage and raises the bar on the compressive capabilities of latent reasoning models. Experiments on the Llama 3.2 1B model and the GSM8K-Aug dataset show a $52\%$ drop in total reasoning length with no penalty to accuracy. In future work, we plan to extend to additional models and datasets, analyze relationships between training coefficients, experiment with architecture variations, and continue our knowledge distillation for latent reasoning SFT efforts. We make our code and pretrained weights available at \url{https://github.com/apning/adaptive-latent-reasoning}.

\end{abstract}


\section{Introduction}
\label{sec:intro}
Inference time scaling methods aim to increase language model capability by utilizing more compute rather than increasing model size. Search strategies such as Monte Carlo Tree Search (MCTS)~\citep{zhao2023largelanguagemodelscommonsense} or beam search~\citep{xie2023selfevaluationguidedbeamsearch} explore alternative reasoning paths or token sequences during generation, resulting in higher-quality solutions compared to greedy decoding. On the other hand, CoT generates intermediate reasoning steps before producing a final answer, improving performance on difficult tasks. CoT has proven highly effective and has seen major adoption in high-performance language models~\citep{openai2024learningtoreasonwithllms, pichai2024gemini2, deepseekai2025deepseekr1incentivizingreasoningcapability}. Despite the success of CoT, it has also led to issues with \textit{overthinking}, resulting in increased compute and memory usage~\citep{hou2025thinkprunepruninglongchainofthought}. The restriction of CoT reasoning to human language tokens presents additional downsides~\citep{hao2024coconut, shen2025codi}. For one, the constraint to human language itself as the modality for reasoning may be limiting. After all, humans themselves can also perform visuospatial or physical reasoning. Additionally, Transformer-based language models generally consume the same amount of compute per-token. Yet, some tokens (eg. syntax tokens) are much easier to predict than others (eg. an answer to a challenging question), leading to a suboptimal allocation of compute.

Latent reasoning~\citep{hao2024coconut, shen2025codi} Transformer models~\citep{vaswani2023attentionneed} represent a recent advancement in large language models (LLMs) which have made contributions towards increasing reasoning efficiency and shortening reasoning length. Compared to chain-of-thought (CoT) reasoning~\citep{wei2023chainofthoughtpromptingelicitsreasoning, kojima2023largelanguagemodelszeroshot}, latent reasoning (alternatively "latent thinking" or "continuous thinking", among other variants) directly uses a Transformer model's previous output hidden state as the next input, instead of using vocabulary tokens to compose a reasoning sequence. \cite{hao2024coconut} and more recently~\cite{shen2025codi} devise methods to develop latent reasoning ability through supervised fine-tuning (SFT), achieving significant reduction in reasoning length compared to CoT, albeit with lower accuracy on certain tasks. However, these methods also primarily employ a \textit{pre-determined} latent reasoning length; that is, the model cannot, on its own, determine when to stop reasoning\footnote{\cite{hao2024coconut} mentioned experimenting with a binary classifier to enable the model to determine when to stop reasoning. They said that this worked similarly well as using a constant reasoning length, and thus, they used the latter for their experiments. They did not provide further details on the former.}. The ability for the model to determine, on its own, when to terminate reasoning (adaptive latent reasoning) presents potential advantages. Not all problems require the same amount of reasoning. Thus, a model with adaptive latent reasoning ability can use fewer thoughts on easier tasks, saving compute and time, while simultaneously thinking longer on harder tasks, potentially improving accuracy on challenging problems. In this work, we explore training latent reasoning LLMs through both SFT and reinforcement learning (RL) which have the ability to decide when to stop their reasoning process and adapt their reasoning lengths based on the input. Additionally, we present unsuccessful knowledge distillation for latent reasoning SFT experiments in Appendix~\ref{sec:sft_knowledge_distill}.

\section{Background}
\label{sec:background}

\begin{figure}[H]
    \centering
    \setlength{\fboxsep}{0pt}
    \includegraphics[width=\textwidth]{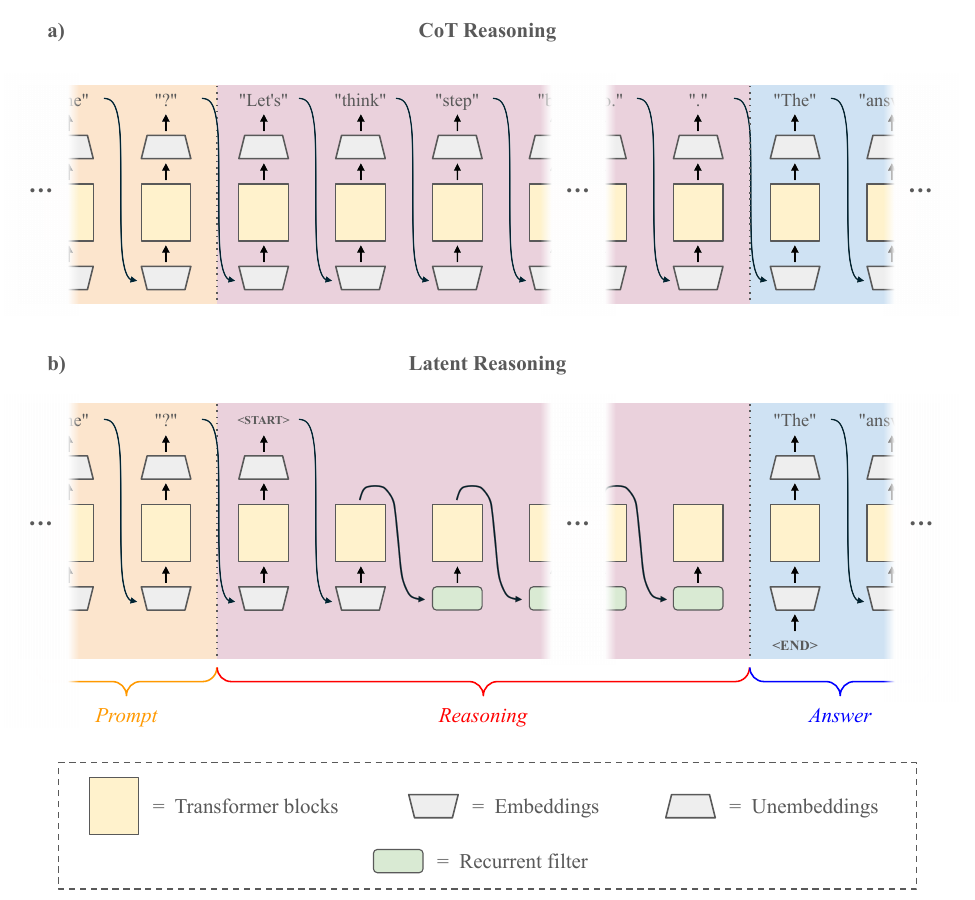}
   \caption{Schematic illustrating the operation of latent reasoning compared to CoT reasoning. Latent reasoning architecture shown reflects the methodology used by this work. Variations may exist in other works, including but not limited to in the use of the special tokens (here \texttt{<START>} and \texttt{<END>}) as well as the recurrent filter.}
    \label{fig:cot_vs_latent}
\end{figure}

\subsection{Reasoning Length Reduction via Reinforcement Learning}

To address the issue of overthinking, reinforcement learning has been employed to reduce reasoning lengths in CoT models. During the training process of Kimi k1.5~\citep{kimiteam2025kimik15scalingreinforcement}, a length penalty was used which penalized the longer responses within a group of generations to the same input. THINKPRUNE~\citep{hou2025thinkprunepruninglongchainofthought} utilized a simple methodology of clipping generations beyond a maximum length, which resulted in the model learning to generate shorter sequences as clipped generations usually could not have an answer extracted, and thus incurred no reward. The maximum length was then subject to an schedule of iterative reductions, allowing the model to gradually learn to produce shorter reasoning sequences with minimal impacts on performance. 

\subsection{Latent Reasoning}

Figure~\ref{fig:cot_vs_latent} visualizes the structure of latent reasoning compared to CoT reasoning. In latent reasoning, the final latent state of the Transformer is passed back in as input. In some formulations, the latent state is passed through a module denoted here as the \textit{recurrent filter} before being input into the Transformer. For example, \cite{shen2025codi} uses a two-layer MLP followed by a layernorm as the recurrent filter. Through the direct passage of the information-rich latent states, capable of simultaneously representing many parallel ideas, latent reasoning implements breadth-first-search-like computations and has been able to achieve impressive compression of reasoning length compared with standard CoT~\citep{hao2024coconut, shen2025codi}.

\subsection{Supervised Fine-Tuning for Latent Reasoning}

COCONUT~\citep{hao2024coconut} and CODI~\citep{shen2025codi} both employ supervised fine-tuning (SFT) on pretrained language models to develop latent reasoning capability. In both cases, large QA-style datasets with CoT reasoning sequences are used. COCONUT uses a staged curriculum learning strategy where increasingly large proportions of the CoT reasoning sequence are replaced with latent reasoning during training. Meanwhile, CODI takes a self-distillation approach. For a given question and answer pair, one forward pass is made using the CoT reasoning sequence (the "teacher") while a separate forward pass is made using a latent reasoning sequence (the "student"). Cross-entropy (CE) next-token-prediction loss is applied to the CoT reasoning sequence tokens and answer tokens of the teacher pass, forming the loss component $\mathcal{L}_\text{teacher}$. Similarly, CE loss is applied to the answer tokens of the student pass, forming the loss component $\mathcal{L}_\text{student}$. CE loss cannot be applied to latent reasoning tokens because there are no labels for latent reasoning. Finally, the knowledge distillation loss component, $\mathcal{L}_\text{KD}$, constrains the latent states of the token directly preceding the answer tokens in the student pass to the latent states of the same token in the teacher pass. This is motivated by the observation that the latent states of this token are rich in reasoning information. Effectively, CODI works by simultaneously training the model on CoT reasoning while also self-distilling that knowledge to latent reasoning. According to the results of \cite{shen2025codi}, CODI trained models outperform their COCONUT counterparts on all tested models and tasks.

\section{Methodology}
\label{sec:methodology}

\subsection{Model Architecture}

We utilize a $2$-layer MLP followed by a layernorm as the recurrent filter. This is identical to that used by CODI. Also similar to CODI, we use special tokens \texttt{<START>} and \texttt{<END>} to denote the beginning and end of latent reasoning sequences. In order to enable adaptive latent reasoning lengths, we add a simple linear binary classification head to the end of the model. During latent reasoning, the binary head indicates at each step whether the model should continue reasoning or stop. 

\subsection{Supervised Fine-Tuning}
\label{subsec:sft}

We use the Llama 3.2 1B Instruct model~\citep{meta2024llama321b}. Following other recent reasoning work~\citep{hao2024coconut, shen2025codi, wu2025parallelcontinuouschainofthoughtjacobi}, we use the GSM8K-Aug dataset~\citep{whynlp2025gsm8kaug, deng2023implicitchainthoughtreasoning}, which is a variant of GSM8K~\citep{cobbe2021trainingverifierssolvemath} extended to $385$k rows via GPT-4 prompting. Out of $385620$ samples in the GSM8K-Aug training set, we find that $194$ have answers that are not plain numbers and filter them out. We utilize the CODI methodology to train our model during the SFT phase. Following in CODI's footsteps, we remove rows from the GSM8K-Aug training dataset which contain negative-valued answers in order to prevent knowledge distillation issues. This removes another $1.6$k rows. Finally, we set aside $15\%$ of the training dataset for the downstream RL phase, leaving $326$k rows for training. We find that this $15\%$ reduction in SFT training data leads to a $\sim 3\text{-}5\%$ reduction in post-SFT test accuracy.

Each row of GSM8K-Aug is composed of a question, a discrete number of separate reasoning steps, and the answer. During SFT, we vary the number of latent reasoning steps per training sample according to the number of reasoning steps belonging to that sample, $k$, a constant multiplier $c$, a constant additive bias $b$, an optional minimum value $n_\text{min}$, and an optional maximum value $n_\text{max}$. For a training sample with $k$ reasoning steps, we therefore train the model with a number of latent reasoning steps according to the equation:

$$
\text{Latent reasoning steps} = 
\min\bigl(n_\text{max}, \max(n_\text{min}, c \cdot k + b)\bigr)
$$

Which simply multiplies $k$ by $c$, adds $b$, then clamps the values between $n_\text{min}$ and $n_\text{max}$.

We use LoRA~\citep{hu2021loralowrankadaptationlarge} to train during both SFT and RL phases.

\subsection{Reinforcement Learning}

Following SFT, we use reinforcement learning to refine the latent reasoning model with the following goals:

\begin{enumerate}
    \item Reduce reasoning lengths on easy questions while maintaining accuracy.
    \item Under the assumption that \textit{more} reasoning is \textit{better}, increase reasoning lengths on more difficult questions to improve accuracy.
\end{enumerate}

We utilize Group Relative Policy Optimization (GRPO)~\citep{shao2024deepseekmathpushinglimitsmathematical}, a recent on-policy RL algorithm which has proven popular for reasoning works~\citep{deepseekai2025deepseekr1incentivizingreasoningcapability, hou2025thinkprunepruninglongchainofthought, yang2024qwen25mathtechnicalreportmathematical, acemath2024}. GRPO is a variant of PPO~\citep{schulman2017proximalpolicyoptimizationalgorithms} which eliminates the critic and works by sampling multiple outputs for the same prompt (the "group") and then calculating the advantages by normalizing rewards within the group. Beyond being generally effective, the group-relative mechanism of GRPO also turns out to perfectly suit our reward function.

We utilize a simple correctness reward which assigns $1$ reward for correctness and $0$ otherwise. Formally, given a question $x$, its answer label $y$, and a sampled answer $\hat{y}$, this is:

\begin{equation}
\text{correct\_reward}(y, \hat{y}) = \begin{cases}
    1 &\text{if } \hat{y} == y \\
    0 & \text{else}
\end{cases}
\end{equation}

We also define a formatting penalty, $\text{format\_penalty}(\hat{y}_i)$, which subtracts $1$ reward if the output $\hat{y}_i$ does not contain the required answer prefix. If an output does not meet the formatting requirements, the answer is automatically considered incorrect, since without the answer prefix the answer cannot be properly extracted.

\begin{equation}
\text{format\_penalty}(\hat{y}) = \begin{cases}
    0 &\text{if } \hat{y} \text{ is properly formatted} \\
    -1 & \text{else}
\end{cases}
\end{equation}

In meeting our first goal of RL, we modify the length penalty used during the development of Kimi k1.5~\citep{kimiteam2025kimik15scalingreinforcement} by only applying the penalty to correct responses and by centering the length penalty so it has a mean of $0$. We term it the \textit{relative length penalty}. For a question $x$ and its answer label $y$, denote the \textit{group} of sampled answers for $x$ as $\{\hat{y}_i\}$ and the corresponding group of reasoning lengths as $\{l_i\}$. Considering only the set of reasoning lengths corresponding to \textit{correct} answers, we denote the mean reasoning length as $\text{mean\_len}$, the minimum reasoning length as $\text{min\_len}$, and the maximum length as $\text{max\_len}$. If all correct responses in a group have the same reasoning length, we assign a relative length penalty of $0$ to all of them. Otherwise, given a relative length penalty coefficient $\lambda_\text{penalty}$, we define the relative length penalty as:

\begin{equation}
\text{rel\_len\_penalty}(y, \hat{y}_i, l_i) = \begin{cases}
    -\lambda_\text{penalty} \frac{l_i - \text{mean\_len}}{\text{max\_len} - \text{min\_len}} &\text{ if } \hat{y}_i == y\\
    0 &\text{ else}
\end{cases}
\end{equation}

We only apply the relative length penalty to \textit{correct} responses in order to avoid discouraging the model from trying harder on more challenging inputs. Additionally, we mean center the relative length penalty around $0$ to prevent reward hacking where the model trades accuracy for shorter reasoning length; the only way for the model to increase the \textit{overall} reward across a group is to improve its accuracy. Despite these efforts, we find that the relative length penalty by itself can still encourage the model to reduce reasoning lengths at the cost of accuracy. As a result, we introduce a relative length \textit{reward} to fulfill our second goal of RL.

We define the relative length reward as exactly the negative of the relative length penalty. That is, given $x$, $y$, $\{y_i\}$, $\{l_i\}$, $\text{mean\_len}$, $\text{max\_len}$, and $\text{min\_len}$ as defined previously, if all correct responses in a group have the same reasoning length, then the relative length reward for all responses will be $0$. Otherwise, given a relative length \textit{reward} coefficient $\lambda_\text{reward}$, we define it as:

\begin{equation}
\text{rel\_len\_reward}(y, \hat{y}_i, l_i) = \begin{cases}
    \lambda_\text{reward} \frac{l_i - \text{mean\_len}}{\text{max\_len} - \text{min\_len}} &\text{ if } \hat{y}_i\ == y\\
    0 &\text{ else}
\end{cases}
\end{equation}

Of course, if applied together, the relative length penalty and relative length reward would simplify into a cancellation of the terms $\lambda_\text{penalty}$ and $\lambda_\text{reward}$. The purpose of the relative length reward is to encourage \textit{longer} reasoning lengths on more difficult questions. We make the assumption that if \textit{some} but not \textit{all} of the answers in a group are correct, then the question should be harder than one in which \textit{all} answers in the group are correct. As a result, we utilize the proportion of correct answers within a group to gauge the difficulty of a question. That is, for a set of sampled answers $\{\hat{y}_i\}$, $p_\text{correct}(\{\hat{y}_i\}) = \frac{\text{n\_correct}(\{\hat{y}_i\})}{|\{\hat{y}_i\}|}$. Then, we set a coefficient $p_\text{cutoff}$ and apply the relative length penalty if the proportion of correct answers within the group is at or above $p_\text{cutoff}$ and apply the relative length reward otherwise.

Therefore, our final reward function becomes:

\begin{equation}
\begin{aligned}
\text{reward}(y, \hat{y}_i, l_i)
    &= \text{correct\_reward}(y, \hat{y}_i)
     + \text{format\_penalty}(\hat{y}_i) \\
    &\quad + \begin{cases}
        \text{rel\_len\_penalty}(y, \hat{y}_i, l_i) &\text{ if } p_\text{correct}(\{\hat{y}_i\}) \geq p_\text{cutoff},\\
        \text{rel\_len\_reward}(y, \hat{y}_i, l_i)  &\text{ otherwise}
    \end{cases}
\end{aligned}
\end{equation}


\subsection{Evaluation}

We use greedy decoding during evaluation. To evaluate GSM8K-Aug correctness, we use regex to extract the number from the answer and then compare to the label with a small numerical tolerance.

\section{Results \& Discussion}
\label{sec:results_and_discussion}

\begin{table}[H]
    \centering
    \renewcommand{\arraystretch}{1.3}
    \begin{tabular}{c|c|c|c|c|c|c|c}
        \hline
        Model & Accuracy & Avg Tok. & Min Tok. & Max Tok. & Tok. Std & Tok. Change & Compress. Ratio\\
        \hline
        \texttt{CoT SFT} & $\mathbf{57.09\%}$ & $28.54$ & $0$ & $97$ & $12.70$ & & 1\\
        \texttt{No-CoT SFT} & $26.46\%$ & $0$ & $0$ & $0$ & $0$ & \\
        \texttt{Latent-6} & $49.73\%$ & $8$ & $8$ & $8$ & $0$ & & $3.57$\\
        \texttt{Latent-6 + RL} & $50.11\%$ & $\mathbf{3.76}$ & $2$ & $8$ & $1.90$ & $-52.94\%$ & $\mathbf{7.58}$\\
        \texttt{Latent-6-by-1} & $51.40\%$ & $10.28$ & $8$ & $13$ & $1.10$ &  & $2.78$\\
        \texttt{Latent-6-by-1 + RL} & $48.14\%$ & $3.91$ & $2$ & $12$ & $2.71$ & $\mathbf{-61.99\%}$ & $7.31$\\
        \hline
        
    \end{tabular}
    \vspace{5mm}
    \caption{Results on GSM8K-Aug test set. "Tok." refers to the number of reasoning tokens used. "Tok. Change" denotes the change in reasoning tokens after RL. The value "Compress. Ratio" for a given model refers to the ratio of the Avg Tok. of \texttt{CoT SFT} over the Avg Tok. of the given model.}
    \label{tab:main_results}
\end{table}

We train the Llama 3.2 1B model into the following variants:

\begin{itemize}
    \item \texttt{CoT SFT}:\quad SFT with normal CoT reasoning.
    \item \texttt{No-CoT SFT}:\quad SFT without CoT reasoning sequence; Answer directly follows question.
    \item \texttt{Latent-6}:\quad SFT to perform latent reasoning; Latent reasoning length is kept to $6$ for all training samples.
    \item \texttt{Latent-6 + RL}:\quad \texttt{Latent-6} model followed by RL training.
    \item \texttt{Latent-6-by-1}:\quad SFT to perform latent reasoning; Latent reasoning step count variables are set as follows: $c = 1$, $b = 6$, and $n_\text{max} = 10$.
    \item \texttt{Latent-6-by-1 + RL}:\quad \texttt{Latent-6-by-1} model followed by RL training.
\end{itemize}

The latent reasoning length of \texttt{Latent-6} is motivated by the results of~\cite{shen2025codi}, which identified a latent reasoning length of $6$ as maximizing accuracy on GSM8K-Aug. Furthermore, we train \texttt{Latent-6-by-1} as the variant with \textit{at least} $6$ latent reasoning steps while also having a variable per-sample latent reasoning length. For fairness of comparison, all variants were trained with only $85\%$ of the training data during SFT, including those which we did not subsequently perform RL on. All RL training was done using the coefficients $\lambda_\text{penalty} = 0.0001$, $\lambda_\text{reward} = 0.1$, and $p_\text{cutoff} = 1$. Also, we note that because of the mandatory special tokens \texttt{<START>} and \texttt{<END>} used to demarcate the beginning and end of latent reasoning, a latent reasoning length of $0$ corresponds to $2$ reasoning \textit{tokens} being used.

We present the results in table~\ref{tab:main_results}. On accuracy, the best latent reasoning model (\texttt{Latent-6-by-1}) scores $5.69\%$ less than \texttt{CoT SFT} but $24.94\%$ better than \texttt{No-CoT SFT}. \texttt{Latent-6-by-1 + RL} achieves the best relative post-RL reduction of reasoning tokens, managing to reduce $61.99\%$ of reasoning tokens (of course, it started out with \textit{more} reasoning tokens than \texttt{Latent-6}). However, it also takes a post-RL accuracy hit of $3.26\%$. In contrast, despite \texttt{Latent-6} scoring $1.67\%$ lower on accuracy than \texttt{Latent-6-by-1}, \texttt{Latent-6 + RL} achieves the lowest reasoning token usage overall while beating \texttt{Latent-6-by-1 + RL} on accuracy by $1.97\%$. Compared to \texttt{Latent-6}, \texttt{Latent-6 + RL} \textit{increases} accuracy by $0.38\%$ while managing to reduce $52.94\%$ of reasoning tokens (which corresponds to a $70.58\%$ reduction in latent reasoning steps).


\begin{figure}[H]
    \centering
    \setlength{\fboxsep}{0pt}
    \includegraphics[width=\textwidth]{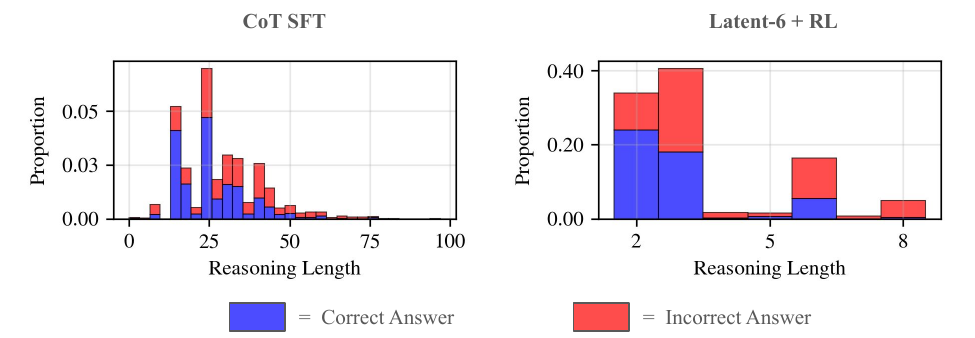}
   \caption{Histograms showing the distributions of reasoning lengths (in tokens) for both \texttt{CoT SFT} and \texttt{Latent-6 + RL} on GSM8K-Aug test set.}
    \label{fig:hist_cot_vs_latent_6_rl}
\end{figure}

\begin{figure}[H]
    \centering
    \setlength{\fboxsep}{0pt}
    \includegraphics[width=\textwidth]{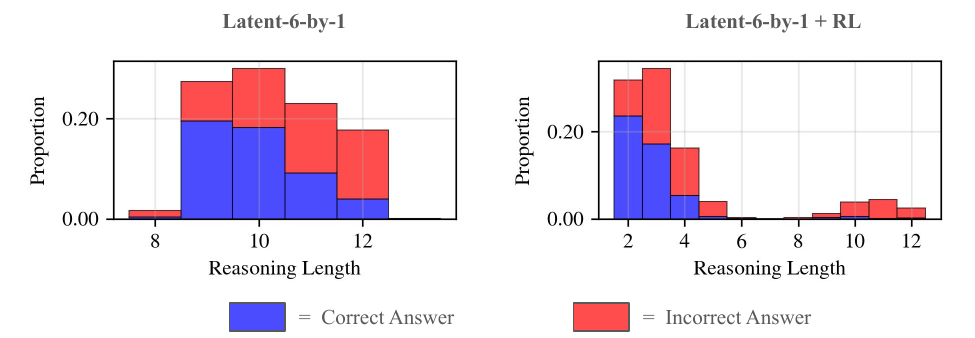}
   \caption{Histograms showing the distributions of reasoning lengths (in tokens) for both \texttt{Latent-6-by-1} and \texttt{Latent-6-by-1 + RL} on GSM8K-Aug test set.}
    \label{fig:hist_latent_6_by_1_vs_rl}
\end{figure}

Figures~\ref{fig:hist_cot_vs_latent_6_rl} and~\ref{fig:hist_latent_6_by_1_vs_rl} show the distribution of reasoning length across the GSM8K-Aug test set for \texttt{CoT SFT}, \texttt{Latent-6 + RL}, \texttt{Latent-6-by-1}, and \texttt{Latent-6-by-1 + RL}. In general, there is a trend where the proportion of incorrect answers increases with reasoning length. This indicates that through both SFT (\texttt{CoT SFT} and \texttt{Latent-6-by-1}) and RL (\texttt{Latent-6 + RL} and \texttt{Latent-6-by-1 + RL}), the model learns to think \textit{more} for harder questions. Table~\ref{tab:main_results} shows that both \texttt{Latent-6 + RL} and \texttt{Latent-6-by-1 + RL} never exceed their respective original pre-RL maximum reasoning lengths. Instead, through Figure~\ref{fig:hist_cot_vs_latent_6_rl} we can see that RL induces the models to decrease the reasoning length used for most samples.

\section{Conclusion and Future Work}

Chain-of-thought (CoT) in language models has enabled more complex reasoning and has experienced great success and adoption. However, the reasoning performed is generally constrained to human language and is often unnecessarily lengthy, leading to increases in compute and memory usage. Latent reasoning has achieved impressive compression of reasoning length over CoT, yet current methods utilize constant, forced latent reasoning lengths. In order to optimize latent reasoning length, we add a binary head to latent reasoning models to enable the model to self-determine when to stop the latent reasoning process. Then, we introduce a post-SFT reinforcement learning training process using GRPO which imposes both relative length penalties and rewards that encourage the model to optimize the latent reasoning length used for each question to minimize length while maintaining accuracy. Utilizing the Llama 3.2 1B model and the GSM8K-Aug dataset, we experiment with a variety of training variants across both SFT and SFT + RL. Through \texttt{Latent-6 + RL}, we manage to achieve a $52.94\%$ post-RL reduction in reasoning tokens with no impact to accuracy. We provide analysis of the various distributions of reasoning token lengths across the test set. 

In future work, we will extend our experimentation to additional models and datasets. We will also analyze the relationships between training coefficients such as $\lambda_\text{penalty}$, $\lambda_\text{reward}$, and $p_\text{cutoff}$ and find more optimal values. Additionally, we will experiment with variations in latent reasoning architecture, such as in the recurrent filter or special tokens. Finally, we will continue to work on our latent reasoning knowledge distillation efforts discussed in Appendix~\ref{sec:sft_knowledge_distill}.

Through the further reduction of latent reasoning length via reinforcement learning, we directly reduce the computational cost and memory usage of latent reasoning Transformer models. Latent reasoning is a growing field which has potential to create more intelligent models while reducing the reasoning lengths currently seen in CoT. By presenting a practical methodology to optimizing adaptive latent reasoning lengths, we are potentially helping to clear the way for future research applying latent reasoning to more challenging tasks where the inefficiencies associated with constant reasoning length may be amplified.

\section{Acknowledgments}

We thank Jose Regio for his insightful comments and suggestions regarding both experimentation and writing. We additionally thank Robert MacKnight and Yusef Ahmed for their assistance with computational systems. This collaboration was facilitated in part by the National Science Foundation Center for Computer Assisted Synthesis (C-CAS) Summer Undergraduate Research Fellowship (SURF), which supported Alex Ning's initial work with the Gomes Group at Carnegie Mellon University. Additionally, Alex Ning thanks Jose Regio for his invaluable mentorship during his SURF.

\bibliographystyle{unsrt}  
\bibliography{references}

\newpage
\appendix

\section{SFT Knowledge Distillation Experiments}
\label{sec:sft_knowledge_distill}

Here we present experiments we performed on knowledge distillation during SFT which we do not consider successful. We present the ideas we explored in subsections \ref{subsec:meaned_reasoning_loss} and \ref{subsec:intermediate_block_loss}. Afterwards, we present the combined results in~\ref{subsec:sft_knowledge_distill_results}.

\subsection{Meaned Reasoning Loss}
\label{subsec:meaned_reasoning_loss}

\begin{figure}[H]
    \centering
    \setlength{\fboxsep}{0pt}
    \includegraphics[width=\textwidth]{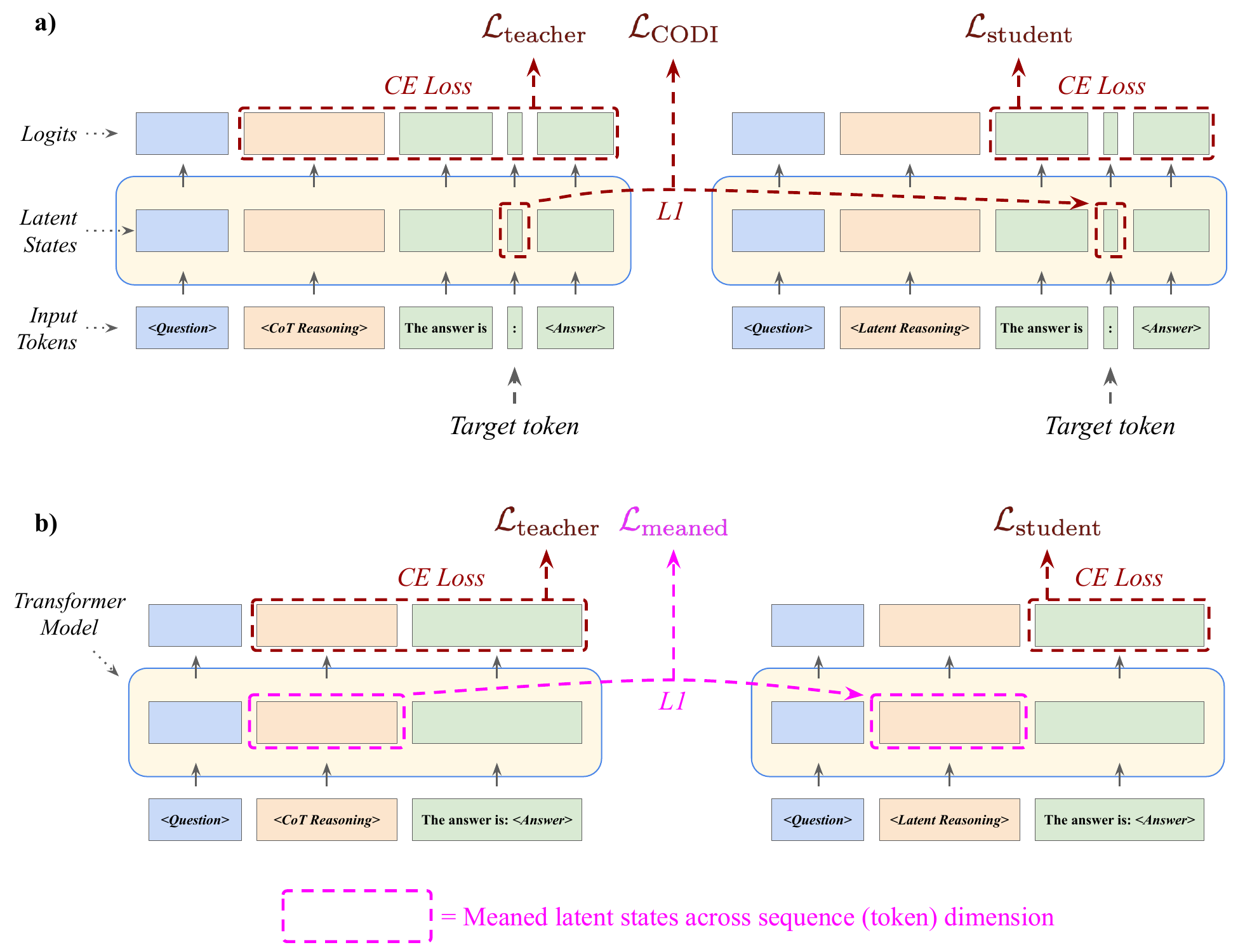}
   \caption{Diagram comparing CODI knowledge distillation (a) with Meaned Reasoning Loss (b). $\mathcal{L}_\text{CODI}$ is the same loss component referred to as $\mathcal{L}_\text{KD}$ by~\cite{shen2025codi}.}
    \label{fig:meaned_vs_codi}
\end{figure}

While the CODI~\citep{shen2025codi} knowledge distillation methodology has resulted in state-of-the-art results, and we employ them in the main results of this paper, the training process can also be characterized as fragile. For example, in order to prevent knowledge distillation issues with CODI on the GSM8K-Aug dataset, the final reasoning step must be removed from all training samples, and all samples with a negative-valued answer must be discarded. Both of these requirements likely stem from CODI's reliance on a single token's latent states to perform knowledge distillation: the final reasoning step in GSM8K-Aug always contains the final answer, which reduces the incentive for the model to gather vital reasoning information into the target token during training~\citep{shen2025codi}. In the case of samples with a negative-valued answer, the minus token ("-") interferes with knowledge distillation (according to comments in the CODI codebase). These constraints present challenges for the practical application of CODI to more advanced reasoning tasks. For example, it may be impractical to avoid reasoning sequences containing the final answer in longer, proof-based mathematical reasoning. Additionally, through the training we have conducted in gathering the results presented throughout this paper, we have found that CODI is often prone to irrecoverable divergence partway through training, which requires re-starting the training process at a checkpoint before the divergence. With that said, we do not present any quantitative metrics to support this claim, as running enough training trials to gather the required data would be very compute expensive. Furthermore, while we have, to the best of our ability, attempted to recreate the original CODI procedure in our own code, we cannot rule out possible differences as the culprit behind these observed divergences.

In order to address the challenges presented by knowledge distillation using a single target token, we present \textit{Meaned Reasoning Loss}. Essentially, Meaned Reasoning Loss distills knowledge between student and teacher outputs in a similar manner to CODI \textit{except} that it uses the \textit{meaned} (ie. averaged) latent states across \textit{all} reasoning tokens. Figure~\ref{fig:meaned_vs_codi} presents a diagram comparing Meaned Reasoning Loss with CODI. Formally, denote $z_{i,j,k}^\text{teacher} \in \R^d$ as the latent state of the $i$-th sample, $j$-th Transformer block, and $k$-th token in the reasoning sequence of the teacher output. Let $z_{i,j,k}^\text{student} \in \R^d$ be similar except that it corresponds to the student output. Denote $I$ as the number of samples, $J$ the number of Transformer blocks, $K_i^\text{teacher}$ the number of reasoning tokens in the teacher sequence for sample $i$, and $K_i^\text{student}$ the number of reasoning tokens in the student sequence for sample $i$. Finally, let $\sigma_j\in \R$ be an optional normalization coefficient corresponding with Transformer block $j$ (ie. normalization by the standard deviation within block $j$ latent states) and $\text{SG}$ be the stop-gradient operation which prevents the knowledge distillation gradient from flowing from the student through the teacher. Then the Meaned Reasoning Loss is:

\begin{equation}
    \mathcal{L}_\text{meaned}
  = \frac{1}{I J} \sum_{i=1}^I \sum_{j=1}^J
    \sigma_j \,
    \text{smooth\_L1}\!\left(
      \frac{1}{K_i^\text{teacher}}\sum_{k=1}^{K_i^\text{teacher}} \text{SG}(z_{i,j,k}^\text{teacher}),
      \frac{1}{K_i^\text{student}}\sum_{k=1}^{K_i^\text{student}} z_{i,j,k}^\text{student}
    \right)
\end{equation}


As in CODI, $\mathcal{L}_\text{meaned}$ would be combined with $\mathcal{L}_\text{teacher}$ and $\mathcal{L}_\text{student}$ - the CE loss of the teacher and student outputs respectively - via a weighted sum to form the final loss $\mathcal{L}$. By relying on the average of all reasoning tokens to distill knowledge, Meaned Reasoning Loss avoids the potential issues associated with knowledge distillation using a single token as medium.

\subsection{Intermediate Block Loss}
\label{subsec:intermediate_block_loss}

\begin{figure}[H]
    \centering
    \setlength{\fboxsep}{0pt}
    \includegraphics[width=\textwidth]{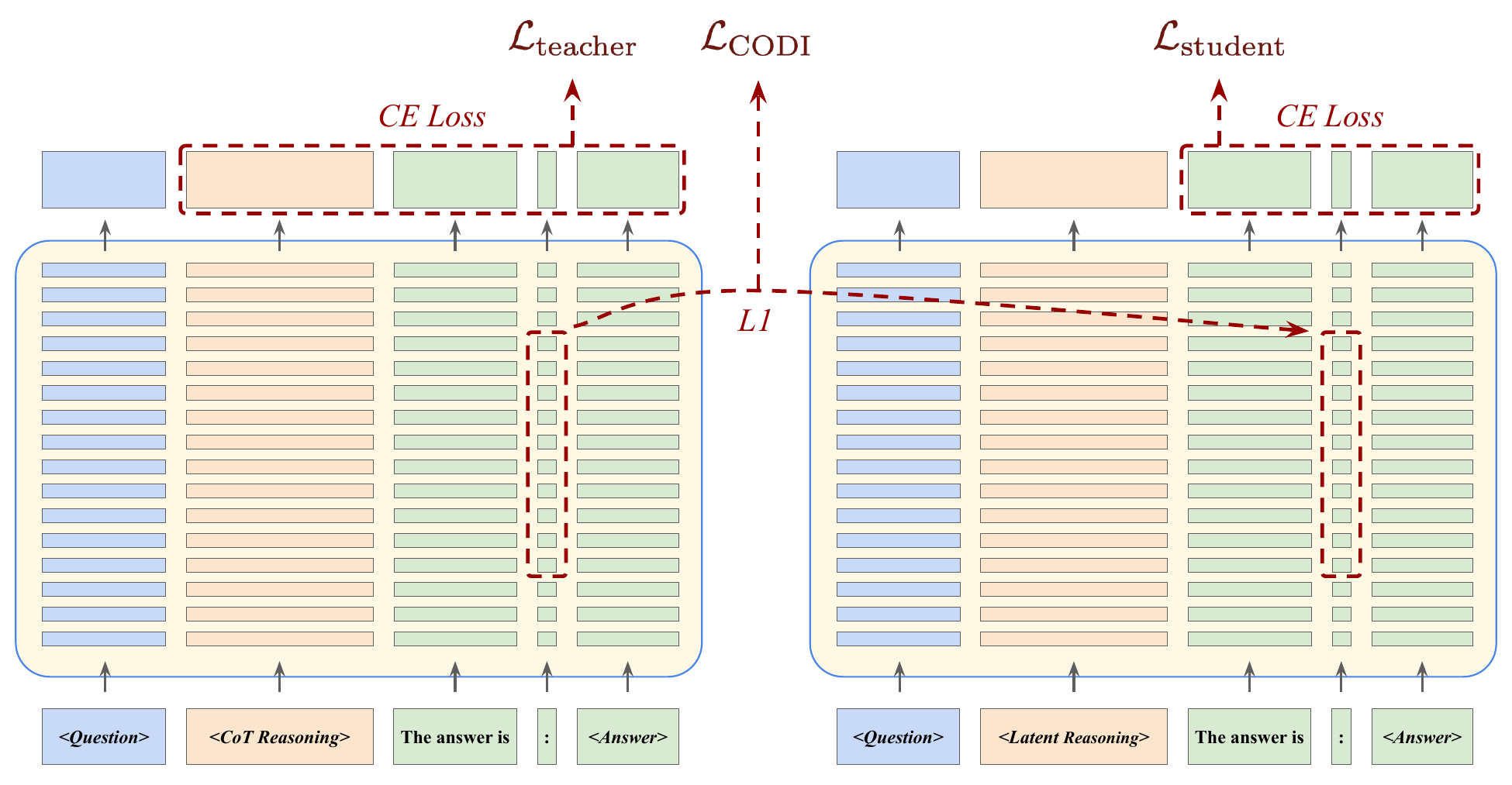}
   \caption{Diagram illustrating Intermediate Block Loss for CODI knowledge distillation. The same concept can similarly be applied to Meaned Reasoning Loss.}
    \label{fig:intermediate_block_loss}
\end{figure}

Although the different blocks of a Transformer are generally structurally identical, they play very different roles. Earlier Transformer blocks resolve discrete tokens into actionable intermediate representations~\citep{nanda2023factfinding}, intermediate Transformer blocks - which share a common understanding of intermediate feature representations - operate on these features, constructing and refining ideas. Finally, the last Transformer blocks heavily denoise the latent states, removing features not relevant to the final task of next-token-prediction~\citep{sun2025transformerlayerspainters, lad2025remarkablerobustnessllmsstages}.

As a result, we hypothesize that performing knowledge distillation using \textit{only} the latent states from the intermediate blocks of the Transformer may present advantages over using all blocks. Latent reasoning neither utilizes discrete tokens as input nor performs next-token-prediction. As such, knowledge distillation which distills teacher reasoning latent states that perform both of these tasks over to the latent reasoning student may result in the student learning a sub-optimal latent reasoning process that contains unnecessary artifacts transferred from CoT reasoning.

Figure~\ref{fig:intermediate_block_loss} illustrates Intermediate Block Loss for CODI knowledge distillation. The loss is only enforced for the latent states of some intermediate Transformer blocks, instead of for the latent states of all blocks of the Transformer. The same idea can be applied analogously to Meaned Reasoning Loss. 

\subsection{Results}
\label{subsec:sft_knowledge_distill_results}

We present results for SFT knowledge distillation variants. The model used is Llama 3.2 1B~\citep{meta2024llama321b} and the dataset is GSM8K-Aug~\citep{whynlp2025gsm8kaug, deng2023implicitchainthoughtreasoning}. Unlike our training procedure mentioned in Section~\ref{subsec:sft}, we use $100\%$ of the training data for SFT since there is no subsequent RL training phase. We filter out a small number of samples with non-numerical answers from the training set as detailed in Section~\ref{subsec:sft}. Following the results of \cite{shen2025codi}, the latent reasoning length utilized is a constant $6$. Llama 3.2 1B has $16$ blocks. For Intermediate Block Loss, we arbitrarily define the intermediate blocks as the middle $10$ blocks, excluding the first $3$ blocks and final $3$ blocks. The following variants are trained:

\begin{itemize}
    \item \texttt{codi}: Model trained with CODI. The last reasoning step is removed from each training sample. Training samples with negative answers are removed.
    \item \texttt{codi + intermediate}: Model trained with CODI applied to intermediate blocks only. The last reasoning step is removed from each training sample. Training samples with negative answers are removed.
    \item \texttt{meaned}: Model trained with Meaned Reasoning Loss. Training dataset is not further modified.
    \item \texttt{meaned + intermediate}: Model trained with Meaned Reasoning Loss applied to intermediate blocks only. Training dataset is not further modified.
    \item \texttt{meaned + codi}: Model trained with Meaned Reasoning Loss and CODI. The last reasoning step is removed from each training sample. Training samples with negative answers are removed.
\end{itemize}

\begin{figure}[H]
    \centering
    \setlength{\fboxsep}{0pt}
    \includegraphics[width=\textwidth]{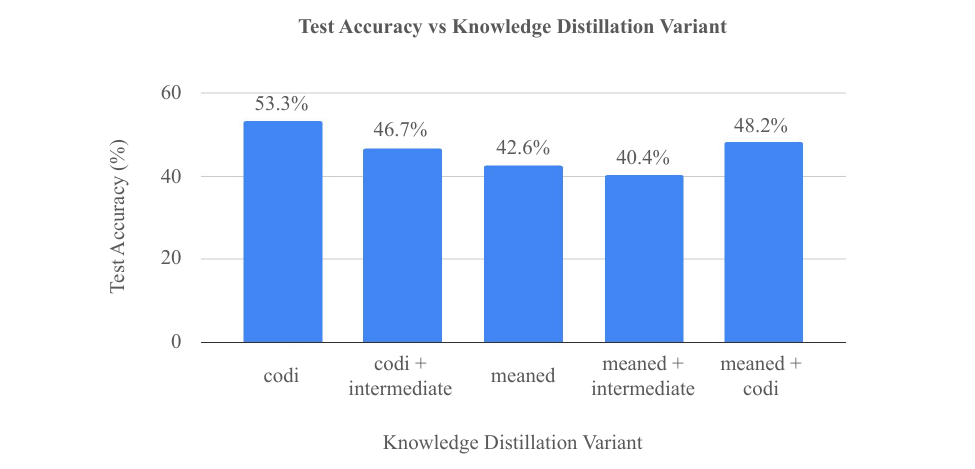}
   \caption{Test accuracy results for each SFT knowledge distillation variant.}
    \label{fig:knowledge_distill_results}
\end{figure}

Figure~\ref{fig:knowledge_distill_results} presents the test accuracy results for each SFT knowledge distillation variant. It can clearly be seen that Meaned Reasoning Loss performs worse than CODI, and that Intermediate Block Loss performs worse when applied to both Meaned Reasoning Loss and CODI. \texttt{meaned + codi} scores better than \texttt{meaned}, but still worse than \texttt{codi} alone. As a result, we conclude that although Meaned Reasoning Loss certainly adds \textit{some} value, performing better than \texttt{No-CoT SFT} from Table~\ref{tab:main_results} in Section~\ref{sec:results_and_discussion} ($42.6\%$ vs. $26.5\%$; although it is not a fair comparison, because \texttt{No-CoT SFT} was trained with only $85\%$ of the training set), both Meaned Reasoning Loss and Intermediate Block Loss, at least in the tested implementations, do not improve test accuracy.

\end{document}